\documentclass[runningheads]{llncs}
\usepackage{graphicx}
\usepackage{amsmath}
\usepackage{amssymb}
\usepackage[toc]{appendix}
\usepackage{algpseudocode}
\usepackage{bm}
\usepackage{dsfont}
\usepackage{nicefrac}
\usepackage{xcolor}
\usepackage{changepage}   

\usepackage{array}
\newcommand{\PreserveBackslash}[1]{\let\temp=\\#1\let\\=\temp}
\newcolumntype{C}[1]{>{\PreserveBackslash\centering}p{#1}}
\newcolumntype{R}[1]{>{\PreserveBackslash\raggedleft}p{#1}}
\newcolumntype{L}[1]{>{\PreserveBackslash\raggedright}p{#1}}

\newcommand\Tstrut{\rule{0pt}{2.3ex}}

\usepackage{tocbasic}
\DeclareNewTOC[%
  type=algo,
  float,
  name=Algorithm,
  listname={List of Algos},
  ]{algo}

\begin{document}
\title{Generating a Graph Colouring Heuristic with Deep Q-Learning and Graph Neural Networks}
%
\titlerunning{Generating a Graph Colouring Heuristic with Deep Q-Learning and GNNs}

\author{George Watkins \and Giovanni Montana \and Juergen Branke}
%
\authorrunning{G. Watkins et al.}
%
\institute{University of Warwick, Coventry, UK \\
\email{george.watkins@warwick.ac.uk, g.montana@warwick.ac.uk} \\
\email{juergen.branke@wbs.ac.uk}} 
\maketitle              
\begin{abstract}

The graph colouring problem consists of assigning labels, or colours, to the vertices of a graph such that no two adjacent vertices share the same colour.  
In this work we investigate whether deep reinforcement learning can be used to discover a competitive construction heuristic for graph colouring. Our proposed approach, ReLCol, uses deep Q-learning together with a graph neural network for feature extraction, and employs a novel way of parameterising the graph that results in improved performance. Using standard benchmark graphs with varied topologies, we empirically evaluate the benefits and limitations of the heuristic learned by ReLCol relative to existing construction algorithms, and demonstrate that reinforcement learning is a promising direction for further research on the graph colouring problem.



\end{abstract}

\section{Introduction}

The Graph Colouring Problem (GCP) is among the most well-known and widely studied problems in graph theory \cite{formanowicz2012survey}. Given a graph $G$, a solution to GCP is an assignment of colours to vertices such that adjacent vertices have different colours; the objective is to find an assignment that uses the minimum number of colours. This value is called the \textit{chromatic number} of $G$, and denoted $\chi(G)$. 
GCP is one of the most important and relevant problems in discrete mathematics, with wide-ranging applications from trivial tasks like sudoku through to vital logistical challenges like scheduling and frequency assignment \cite{ahmed2012applications}.
Given that GCP has been proven to be NP-Complete for general graphs \cite{garey1979computers}, no method currently exists that can optimally colour any graph in polynomial time.
Indeed it is hard to find even approximate solutions to GCP efficiently \cite{lund1994hardness} and currently no algorithm with reasonable performance guarantees exists \cite{korte2011combinatorial}.

Many existing methods for GCP fall into the category of \textit{construction heuristics}, which build a solution incrementally. Designing an effective construction heuristic is challenging and time-consuming and thus there has been a lot of interest in ways to generate heuristics \textit{automatically}. This has previously been very successful in, for example, job shop scheduling \cite{branke2015automated}. Among the simplest construction methods for GCP are \textit{greedy algorithms} \cite{lewis2015guide}, in which vertices are selected one by one and assigned the `lowest' permissible colour based on some pre-defined ordering of colours. 





In this work we
investigate the use of reinforcement learning (RL) to learn a greedy construction heuristic for GCP by framing the selection of vertices as a sequential decision-making problem. 
Our proposed algorithm, ReLCol, uses deep Q-learning (DQN) \cite{mnih2013playing} together with a graph neural network (GNN) \cite{scarselli2008graph,battaglia2018relational} to learn a policy that selects the vertices for our greedy algorithm. 
Using existing benchmark graphs, we compare the performance of the ReLCol heuristic against several existing greedy algorithms, notably Largest First, Smallest Last and DSATUR. Our results indicate that the solutions generated by our heuristic are competitive with, and in some cases better than, these methods. As part of ReLCol, we also present an alternative way of parameterising the graph within the GNN, and show that our approach significantly improves performance compared to the standard representation.






\section{Related Work}




\subsubsection{Graph Colouring}

Methods for GCP, as with other combinatorial optimisation (CO) problems, can be separated into exact solvers and heuristic methods. Exact solvers must process an exponentially large number of solutions to guarantee optimality; as such, they quickly become computationally intractable as the size of the problem grows \cite{de2018exact}. Indeed exact algorithms are generally not able to solve GCP in reasonable time when the number of vertices exceeds 100 \cite{moalic2018variations}. 

When assurances of optimality are not required, heuristic methods offer a compromise between good-quality solutions and reasonable computation time. Heuristics may in some cases produce optimal solutions, but offer no guarantees for general graphs. 
Considering their simplicity,
greedy algorithms are very effective: even ordering the vertices at random can yield a good solution. And crucially, for every graph there exists a vertex sequence such that greedily colouring the vertices in that order will yield an optimal colouring \cite{lewis2015guide}. 

Largest-First (LF), Smallest-Last (SL) and DSATUR \cite{brelaz1979new} are the three most popular such algorithms \cite{janczewski2001smallest}, among which DSATUR has become the de facto standard for GCP \cite{sager1991pruning}. As such, we have chosen these three heuristics as the basis for our comparisons.
Both LF and SL are \textit{static} methods, meaning the vertex order they yield is fixed at the outset. LF chooses the vertices in decreasing order by degree; SL also uses degrees, but selects the vertex $v$ with smallest degree to go last, and then repeats this process with the vertex $v$ (and all its incident edges) removed. Conversely, DSATUR is a \textit{dynamic} algorithm: at a given moment the choice of vertex depends on the previously coloured vertices. DSATUR selects the vertex with maximum \textit{saturation}, where saturation is the number of distinct colours assigned to its neighbours. Similarly, the Recursive Largest First algorithm \cite{leighton1979graph} is dynamic. At each step it finds a maximal independent set and assigns the same colour to the constituent vertices. The coloured vertices are then removed from the graph and the process repeats. 

\textit{Improvement algorithms} take a different approach: given a (possibly invalid) colour assignment, these methods use local search to make small adjustments in an effort to improve the colouring, either by reducing the number of colours used or eliminating conflicts between adjacent vertices. Examples include TabuCol \cite{hertz1987using}, simulated annealing \cite{aragon1991optimization} and evolutionary algorithms \cite{galinier1999hybrid}. 

\subsubsection{Machine Learning methods for CO problems}
Given how difficult it is to solve CO problems exactly, and the reliance on heuristics for computationally tractable methods, machine learning appears to be a natural candidate for addressing problems like GCP. Indeed there are many examples of methods for CO problems that use RL  \cite{mazyavkina2021reinforcement,barrett2020exploratory,ireland2022lense} or other machine learning techniques \cite{smith1999neural,bengio2021machine}. 

For GCP, supervised learning has been used to predict the chromatic number of a graph \cite{lemos2019graph} but this is dependent on having a labelled training dataset. Given the computational challenges inherent in finding exact solutions, this imposes limitations on the graphs that can be used for training. Conversely, RL does not require a labelled dataset for training. In \cite{zhou2016reinforcement}, RL is used to support the local search component of a hybrid method for the related $k$-GCP problem by learning the probabilities with which each vertex should be assigned to each colour. While iteratively solving $k$-GCP for decreasing values of $k$ is a valid (and reasonably common) method for solving GCP \cite{lu2010memetic,galinier1999hybrid}, it is inefficient. 




On the other hand, \cite{huang2019coloring} addresses GCP directly with a method inspired by the success of AlphaGo Zero \cite{silver2017mastering}. In contrast to greedy algorithms, this approach uses a pre-determined vertex order and learns the mechanism for deciding the \textit{colours}. During training they use a computationally demanding Monte Carlo Tree Search using 300 GPUs; due to the computational overhead and lack of available code we were unable to include this algorithm in our study. 

Our proposed algorithm is most closely related to \cite{gianinazzi2021learning}, in which the authors present a greedy construction heuristic that uses RL with an attention mechanism \cite{vaswani2017attention} to select the vertices. There are, however, several key differences between the two methods. Their approach uses the REINFORCE algorithm \cite{williams1992simple} whereas we choose DQN \cite{mnih2013playing} because the action space is discrete; they incorporate spatial and temporal locality biases; 
and finally, we use a novel state parameterisation, which we show improves the performance of our algorithm. 




\section{Problem Definition}
A \textit{$k$-colouring} of a graph $G = (V,E)$ is a partition of the vertices $V$ into $k$ disjoint subsets such that, for any edge $(u,v) \in E$, the vertices $u$ and $v$ are in different subsets. The subsets are typically referred to as \textit{colours}. GCP then consists of identifying, for a given graph $G$, the minimum number of colours for which a $k$-colouring exists and the corresponding colour assignment. This number is known as the \textit{chromatic number} of $G$, denoted $\chi(G)$. 

Given any graph, a greedy construction heuristic determines the order in which vertices are to be coloured, sequentially assigning to them the lowest permissible colour according to some pre-defined ordering of colours. In this work we address the problem of automatically deriving a greedy construction heuristic that colours general graphs using as few colours as possible. 





\section{Preliminaries}

\subsubsection{Markov Decision Processes}
A Markov Decision Process (MDP) is a discrete-time stochastic process for modelling the decisions taken by an agent in an environment. 
An MDP is specified by the tuple $(\mathcal{S}, \mathcal{A}, \mathcal{P}, \mathcal{R}, \gamma)$, where $\mathcal{S}$ and $\mathcal{A}$ represent the state and action spaces; $\mathcal{P}$ describes the environment's transition dynamics; $\mathcal{R}$ is the reward function; and $\gamma$ is the discount factor. 
The goal of reinforcement learning is to learn a decision policy $\pi:\mathcal{S} \rightarrow \mathcal{A}$ that maximises the expected sum of discounted rewards, $\mathop{\mathbb{E}} \left[ \sum_{t=1}^{\infty} \gamma^t R_{t}\right]$.

\subsubsection{Deep Q-Learning}
Q-learning \cite{watkins1992q} is a model-free RL algorithm that learns $Q^{*}(s,a)$, the value of taking action $a$ in a state $s$ and subsequently behaving optimally. Known as the \textit{optimal action value function}, $Q^{*}(s,a)$ is defined as
\begin{equation} \label{eq:q_func}
Q^{*}(s,a) = \mathop{\mathbb{E}} \left[ \sum_{i=1}^{\infty} \gamma^i R_{t+i} \;\middle|\; S_t=s, A_t=a \right]
\end{equation}
where $S_t, A_t$ and $R_t$ are random variables representing respectively the state, action and reward at timestep $t$. \textit{Deep} Q-learning (DQN) \cite{mnih2013playing} employs a Q-network
parameterised by weights $\theta$ to approximate $Q^{*}(s,a)$.
Actions are chosen greedily with respect to their values with probability $1-\epsilon$, and a random action is taken otherwise to facilitate exploration.
Transitions $(s,a,r,s^{\prime})$ - respectively the state, action, reward and next state - are added to a buffer and the Q-network is trained by randomly sampling transitions, backpropagating the loss 
\begin{equation} \label{equation:loss}
L(\theta) = \left( \left[ r + \gamma \max_{a^{\prime}}Q_{\hat{\theta}}(s^{\prime}, a^{\prime}) \right] - Q_{\theta}(s,a) \right) ^2
\end{equation}
and updating the weights using stochastic gradient descent.
Here $Q_{\theta}(s,a)$ and $Q_{\hat{\theta}}(s,a)$ are estimates of the value of state-action pair $(s,a)$ using the Q-network and a \textit{target network} respectively. The target network is a copy of the Q-network, with weights that are updated via periodic soft updates, $\hat{\theta} \leftarrow \tau \theta + (1-\tau) \hat{\theta}$. Using the target network in the loss helps to stabilise learning \cite{mnih2013playing}. 

\subsubsection{Graph Neural Networks}

\begin{figure}[t]
\begin{center}
\includegraphics[width=12cm]{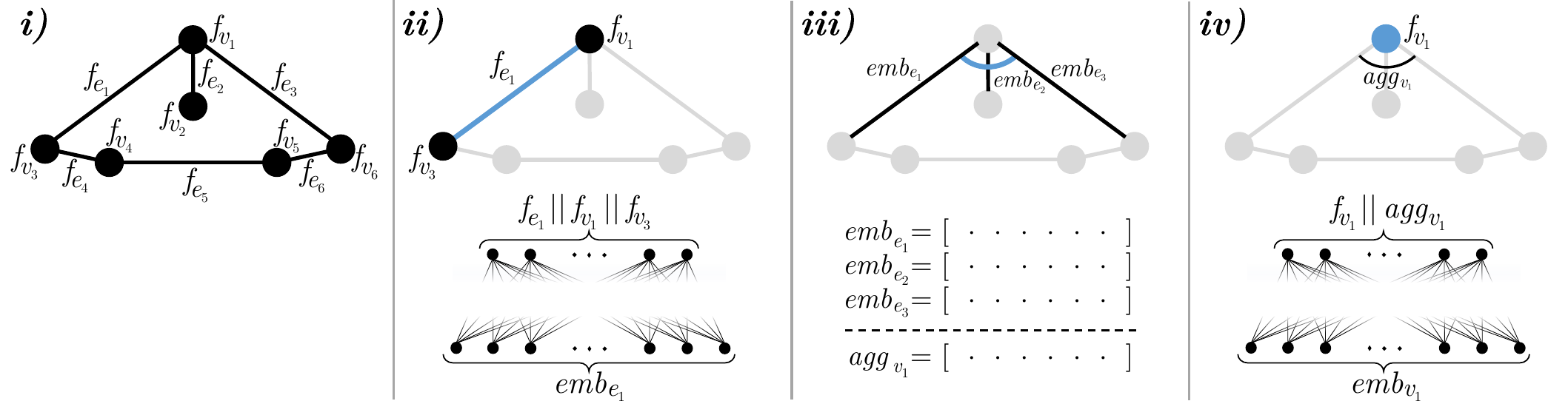}
\caption{Demonstration of how a GNN block generates edge and vertex embeddings. Blue indicates the element that is updated in that step. \textit{\textbf{i)}} The input graph with edge and vertex features; \textit{\textbf{ii)}} For each edge $e$, concatenate the features of $e$ with the features of the vertices it connects, and pass the resulting vector through a small neural network to generate the edge embedding $emb_{e}$; \textit{\textbf{iii)}} For each vertex $v$, aggregate the embeddings of the incident edges using an elementwise operation like \textit{sum} or \textit{max} to generate the edge aggregation $agg_{v}$; \textit{\textbf{iv)}} For each vertex $v$, concatenate the features of $v$ with the associated edge aggregation $agg_{v}$, and pass the resulting vector through a small neural network to generate the vertex embedding $emb_{v}$. Blocks can be stacked by repeating this process with the previous block's edge and vertex embeddings used as the features.}
\label{fig:gnn_demo}
\end{center}
\end{figure}

Graph neural networks (GNNs) \cite{scarselli2008graph,battaglia2018relational} support learning over graph-structured data. GNNs consist of blocks; the most general GNN block takes a graph $G$ with vertex-, edge- and graph-level features, and outputs a new graph $G^{\prime}$ with the same topology as $G$ but with the features replaced by vertex-, edge- and graph-level embeddings \cite{battaglia2018relational}. The embeddings are generated via a message-passing and aggregation mechanism whereby information flows between pairs of neighbouring vertices. Stacking multiple GNN blocks allows for more complex dependencies to be captured. The steps within a single GNN block are demonstrated in Figure \ref{fig:gnn_demo}; in our method we do not use graph-level features so for simplicity these have been omitted. 

\section{Methodology}

\subsection{Graph colouring as a Markov Decision Process}

\subsubsection{States}
A state $s \in \mathcal{S}$ for graph $G=(V,E)$ is a vertex partition of $V$ into subsets $P_{i}$, $i \in \{-1,0,1,2,...\}$. For $i\neq-1$, the partition $P_{i}$ contains the vertices currently assigned colour $i$, and $P_{-1}$ represents the set of currently un-coloured vertices. States in which $P_{-1} = \emptyset$ are terminal. Our method for parameterising the state, which results in improved performance, is described in Section \ref{section:parameterising_the_state}.

\subsubsection{Actions}
An action $a \in \mathcal{A}$ is an un-coloured vertex (i.e. $a \in P_{-1}$) indicating the next vertex to be coloured. The complete mechanism by which ReLCol chooses actions is described in Section \ref{section:selecting_actions}.

\subsubsection{Transition function}

Given an action $a$, the transition function $\mathcal{P}:\mathcal{S} \times \mathcal{A} \rightarrow \mathcal{S}$ updates the state $s$ of the environment to $s'$ by assigning the lowest permissible colour to vertex $a$. Note that choosing colours in this way does not preclude finding an optimal colouring \cite{lewis2015guide} as every graph admits a sequence that will yield an optimal colouring.
The transition function $\mathcal{P}$ is deterministic: given a state $s$ and an action $a$, there is no uncertainty in the next state $s'$. 


\subsubsection{Reward function}
For GCP, the reward function should encourage the use of fewer colours. As such our reward function for the transition $(s,s')$ 
is defined as
\[
\mathcal{R}(s,s') = -1 (C(s') - C(s))
\]
where $C(s)$ indicates the number of colours used in state $s$.

\subsubsection{Discount factor}
GCP is an episodic task, with an episode corresponding to colouring a single graph $G$. Given that each episode is guaranteed to terminate after $n$ steps, where $n$ is the number of vertices in $G$, we set $\gamma=1$. Using $\gamma < 1$ would bias the heuristic towards deferring the introduction of new colours, which may be undesirable. 





\subsection{Parameterising the state} \label{section:parameterising_the_state}

Recall that for the graph $G=(V,E)$, a state $s \in \mathcal{S}$ is a partition of $V$ into subsets $P_{i}$, $i \in \{-1,0,1,2,...\}$. We represent the state using a \textit{state graph} $G_s = (V, E_s, F^v_s, F^e_s)$: respectively the vertices and edges of $G_s$, together with the associated vertex and edge features.

\subsubsection{State graph vertices}
Note that the vertices in $G_s$ are the same as the vertices in the original graph $G$. Then, given a state $s$, the feature $f^v_s$ of vertex $v$ is a 2-tuple containing: i) A vertex name $\in \{0,1,2,...,n-1\}$ and ii) The current vertex colour $c_v \in \{-1,0,1,2,...\}$ (where $c_v = -1$ if and only if $v$ has not yet been assigned a colour).
\subsubsection{State graph edges}
In the standard GNN implementation, messages are only passed between vertices that are joined by an edge. In our implementation we choose to represent the state as a complete graph on $V$ to allow information to flow between all pairs of vertices. We use a binary edge feature $f^e_s$ to indicate whether the corresponding edge was in the original graph $G$:

\[
f^e_s = 
\begin{cases}
    -1 & \text{if } e=(v_i, v_j) \in E\\
    0 & \text{otherwise}
    \end{cases} 
\]
Our state parameterisation, which is the input to the Q-network, allows messages to be passed between all pairs of vertices, including those that are not connected; in Section \ref{section:complete_graph_results} we show that this representation results in improved performance. 

\subsection{Q-network architecture}

Our Q-network is composed of several stacked GNN blocks followed by a feedforward neural network of fully connected layers with ReLU activations. Within the aggregation steps in the GNN we employ an adaptation of Principal Neighbourhood Aggregation \cite{corso2020principal} which has been shown to mitigate information loss. The GNN takes the state graph $G_s$ as input, and returns a set of vertex embeddings. For each vertex $v$, the corresponding embedding is passed through the fully connected layers to obtain the value for the action of choosing $v$ next. 


\subsection{Selecting actions} \label{section:selecting_actions}




In general actions 
are selected using an $\epsilon$-greedy policy with respect to the vertices' values. However using the Q-network is (relatively) computationally expensive. As such, where it has no negative effect - in terms of the ultimate number of colours used - we employ alternative mechanisms to select vertices.

\subsubsection{First Vertex rule} 
The first vertex to be coloured is selected at random. By Proposition \ref{proposition:first_vertex}, this does not prevent an optimal colouring being found.
\begin{proposition} \label{proposition:first_vertex}
An optimal colouring remains possible regardless of the first vertex to be coloured. \end{proposition}
\textbf{Proof.} Let $G=(V,E)$ be a graph with $\chi(G)=k^*$, and $\mathcal{P^*} = P^*_0, P^*_1, ..., P^*_{k^*-1}$ an optimal colouring of $G$, where the vertices in $P^*_i$ are all assigned colour $i$.
Suppose vertex $v$ is the first to be selected, and $j$ is the colour of $v$ in $\mathcal{P^*}$ (i.e. $v \in P^*_{j}$, where $0 \leq j \leq k^*-1$). Simply swap the labels $P^*_0$ and $P^*_j$ so that the colour assigned to $v$ is the `first' colour.
Now, using this new partition, we can use the construction described in \cite{lewis2015guide} to generate an optimal colouring. \hfill $\blacksquare$




\subsubsection{Isolated Vertices rule} We define a vertex to be \textit{isolated} if all of its neighbours have been coloured. By Proposition \ref{proposition:isolated_vertices}, we can immediately colour any such vertices without affecting the number of colours required. 

\begin{proposition} \label{proposition:isolated_vertices}
Immediately colouring isolated vertices has no effect on the number of colours required to colour the graph.
\end{proposition}
\textbf{Proof.} Let $G=(V,E)$ be a graph with $\chi(G)=k^*$. Suppose also that $\mathcal{P} = P_{-1}, P_0, P_1, ..., P_{k^*-1}$ is a partial colouring of $G$ (with $P_{-1}$ the non-empty set of un-coloured vertices).
Let $v \in P_{-1}$ be an un-coloured, isolated vertex (i.e. it has no neighbours in $P_{-1}$). No matter when $v$ is selected, its colour will be the first that is different from all its neighbours. Also, given that $v$ has no un-coloured neighbours, it has no influence on the colours assigned to subsequent vertices. Therefore $v$ can be chosen at any moment (including immediately) without affecting the ultimate number of colours used. \hfill $\blacksquare$

\begin{figure}[h]
\begin{center}
\includegraphics[width=12cm]{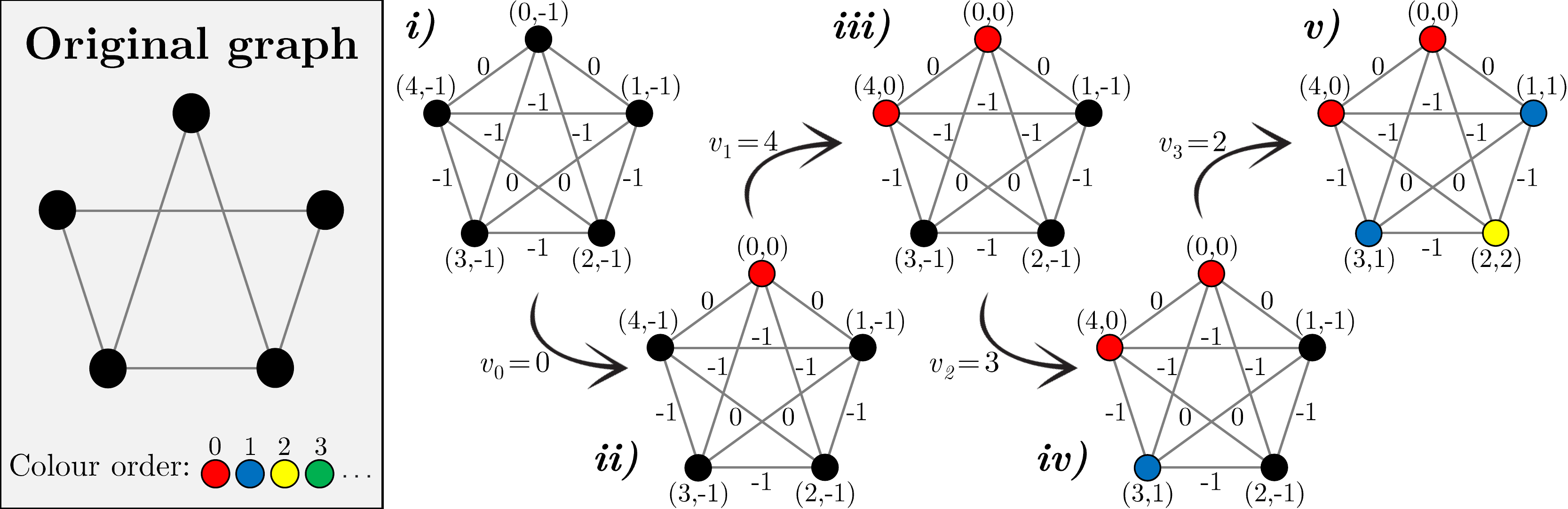}
\caption{Example of graph parameterisation and colouring rules, where $v_t=j$ indicates the vertex with name $j$ is selected at step $t$. \textit{\textbf{i})} Initial parameterisation of the original graph; \textit{\textbf{ii})} First vertex $v_0=0$ is selected at random and assigned colour $0$; \textit{\textbf{iii})} Vertex $v_1=4$ can also be assigned colour $0$; \textit{\textbf{iv})} Vertex $v_2=3$ cannot be assigned colour $0$ so takes colour $1$; \textit{\textbf{v})} Vertex $v_3=2$ cannot be assigned colour $0$ or $1$ so takes colour $2$, leaving vertex $1$ isolated; vertex $1$ cannot be assigned colour $0$ so takes colour $1$.}
\label{fig:example_colouring}
\end{center}
\end{figure}

\subsection{The ReLCol algorithm}

Fig. \ref{fig:example_colouring} demonstrates how the state graph is constructed, and how it evolves as the graph is coloured. The full ReLCol algorithm is presented in Algorithm 1.


\begin{algo} \label{algo:relcol_pseudocode}
\begin{center}
\includegraphics[trim={5cm 13.5cm 5cm 5cm}, width=8.1cm]{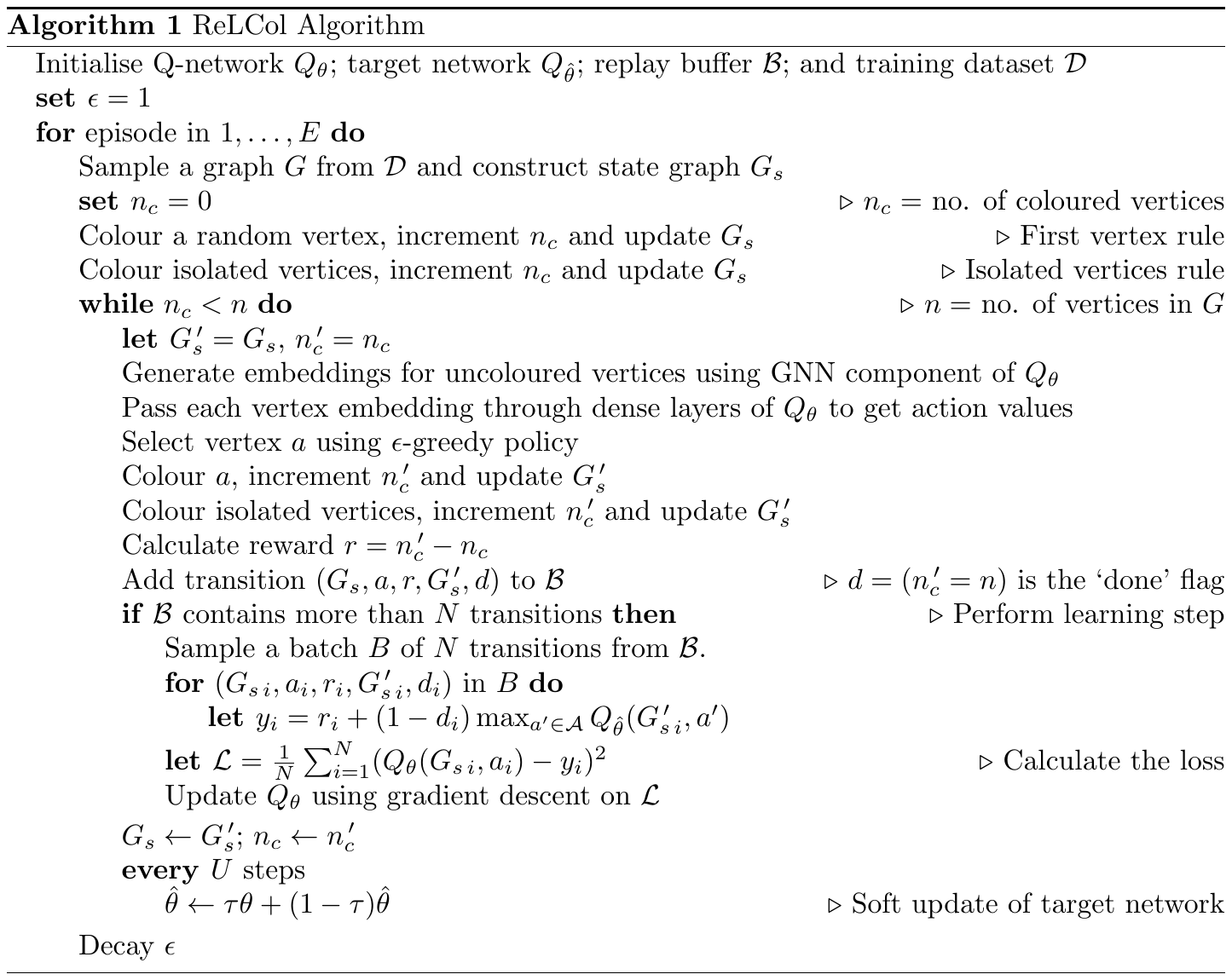}
\end{center}
\end{algo}




\section{Experimental results}

In this section we evaluate the performance of the heuristic learned by ReLCol against existing algorithms. Because the learning process is inherently stochastic, we generate 12 ReLCol heuristics, each using a different random seed. In our experiments we apply all 12 to each graph and report the average number of colours required. We note that although ReLCol refers to the generating algorithm, for brevity we will also refer to our learned heuristics as ReLCol. 


\subsubsection{Architecture and hyperparameters}
Our Q-network consists of 5 GNN blocks and 3 fully connected layers with weights initialised at random. Our GNN blocks use only edge and vertex features; we experimented with including global features but found no evidence of performance improvement. The vertex and edge embeddings, as well as the hidden layers in all fully connected neural networks, have 64 dimensions. We use the Adam optimiser \cite{kingma2014adam} with learning rate 0.001 and batch size 64, $\tau=0.001$, and an $\epsilon$-greedy policy for exploration, where $\epsilon$ decays exponentially from 0.9 to 0.01 through 25000 episodes of training. 

\subsubsection{Training data}

We have constructed a dataset of 1000 training graphs of size $n \in [15,50]$, composed of 7 different graph types: Leighton graphs \cite{leighton1979graph}, Queen graphs \cite{fricke1995combinatorial}, Erdos-Renyi graphs \cite{erdHos1959random}, Watts-Strogatz graphs \cite{watts1998collective}, Barabasi-Albert graphs \cite{barabasi1999emergence}, Gaussian Random Partition graphs \cite{brandes2003experiments} and graphs generated by our own method which constructs graphs with known upper bound on the chromatic number. Each training graph was constructed by choosing its type and size uniformly at random. Where the generating process of the chosen graph type has its own parameters these too were chosen at random to ensure as much diversity as possible amongst the training graphs. Our datasets, together with our code, are available on GitHub\footnote{https://github.com/gpdwatkins/graph\_colouring\_with\_RL}.



\subsection{Comparison with existing algorithms}

We first compare ReLCol to existing construction algorithms, including Largest First, Smallest Last, DSATUR and Random (which selects the vertices in a random order). We also compare to the similar RL-based method presented in Gianinazzi et al. \cite{gianinazzi2021learning}. Using their implementation we generate 12 heuristics with different random seeds and report the average result. Note that in their paper the authors present both a deterministic and a stochastic heuristic, with the stochastic version generated by taking a softmax over the action weights. At test time their stochastic heuristic is run 100 times and the best colouring is returned. Given that with enough attempts even a random algorithm will eventually find an optimal solution, we consider heuristics that return a colouring in a single pass to be more interesting. As such we consider only the deterministic versions of the Gianinazzi et al. algorithm and ReLCol.

Each heuristic is applied to the benchmark graphs used in \cite{lemos2019graph} and \cite{gianinazzi2021learning}, which represent a subset of the graphs specified in the \textit{COLOR02: Graph Colouring and its Generalizations} series\footnote{https://mat.tepper.cmu.edu/COLOR02/}.
For these graphs the chromatic number is known; as such we report the \textit{excess} number of colours used by an algorithm (i.e. 0 would mean the algorithm has found an optimal colouring for a graph).
The results are summarised in Table \ref{table:lemos_comparisons}. On average over all the graphs DSATUR was the best performing algorithm, using 1.2 excess colours, closely followed by our heuristic with 1.35 excess colours. The other tested algorithms perform significantly worse on these graphs, even slightly worse than ordering the vertices at random.
The test set contains a mix of easier graphs - all algorithms manage to find the chromatic number for \texttt{huck} - as well as harder ones - even the best algorithm uses four excess colours on \texttt{queen13\_13}. DSATUR and ReLCol each outperform all other methods on 4 of the graphs.

\begin{table}[h]
\begin{center}
\resizebox{.83\textwidth}{!}{%
    \begin{tabular}{|c C{.55cm} C{.55cm} l C{.8cm} C{.8cm} c l l|} 
     \hline
     \begin{tabular}{@{}c@{}}\textbf{Graph} \\ \textbf{instance} \end{tabular} & $n$ & $\chi$ & \textbf{Random} & \textbf{LF} & \textbf{SL} & \textbf{DSATUR} & \begin{tabular}{@{}c@{}}\textbf{Gianinazzi} \\ \textbf{et al.} \end{tabular} & \textbf{ReLCol} \\
     \hline
    queen5\_5 & 25 & 5 & \hspace{5pt}$2.3^{\pm0.1}$ & 2& 3 & \textbf{0} & \hspace{5pt}$2.1^{\pm0.3}$ & $0.2^{\pm0.11}$ \Tstrut\\
    queen6\_6 & 36 & 7 & \hspace{5pt}$2.4^{\pm0.06}$ & 2& 4 & 2 & \hspace{5pt}$3.2^{\pm0.16}$ & $\bm{1}^{\pm0}$ \\
    myciel5 & 47 & 6 & \hspace{5pt}$0.1^{\pm0.03}$ & 0& 0 & 0 & \hspace{5pt}$0.1^{\pm0.08}$ & $0^{\pm0}$ \\
    queen7\_7 & 49 & 7 & \hspace{5pt}$4^{\pm0.06}$ & 5& 3 & 4 & \hspace{5pt}$3.3^{\pm0.32}$ & $\bm{2.2}^{\pm0.16}$ \\
    queen8\_8 & 64 & 9 & \hspace{5pt}$3.4^{\pm0.07}$ & 4& 5 & 3 & \hspace{5pt}$3.3^{\pm0.24}$ & $\bm{2.1}^{\pm0.14}$ \\
    1-Insertions\_4 & 67 & 4 & \hspace{5pt}$1.2^{\pm0.04}$ & 1& 1 & 1 & \hspace{5pt}$1^{\pm0}$ & $1^{\pm0}$ \\
    huck & 74 & 11 & \hspace{5pt}$0^{\pm0}$ & 0& 0 & 0 & \hspace{5pt}$0^{\pm0}$ & $0^{\pm0}$ \\
    jean & 80 & 10 & \hspace{5pt}$0.3^{\pm0.05}$ & 0& 0 & 0 & \hspace{5pt}$0^{\pm0}$ & $0.3^{\pm0.12}$ \\
    queen9\_9 & 81 & 10 & \hspace{5pt}$3.9^{\pm0.07}$ & 5& 5 & 3 & \hspace{5pt}$5^{\pm0}$ & $\bm{2.7}^{\pm0.25}$ \\
    david & 87 & 11 & \hspace{5pt}$0.7^{\pm0.07}$ & 0& 0 & 0 & \hspace{5pt}$0.4^{\pm0.14}$ & $1.2^{\pm0.33}$ \\
    mug88\_1 & 88 & 4 & \hspace{5pt}$0.1^{\pm0.02}$ & 0& 0 & 0 & \hspace{5pt}$0^{\pm0}$ & $0^{\pm0}$ \\
    myciel6 & 95 & 7 & \hspace{5pt}$0.3^{\pm0.05}$ & 0& 0 & 0 & \hspace{5pt}$0.8^{\pm0.17}$ & $0^{\pm0}$ \\
    queen8\_12 & 96 & 12 & \hspace{5pt}$3.3^{\pm0.06}$ & 3& 3 & \textbf{2} & \hspace{5pt}$4^{\pm0}$ & $2.6^{\pm0.18}$ \\
    games120 & 120 & 9 & \hspace{5pt}$0^{\pm0.01}$ & 0& 0 & 0 & \hspace{5pt}$0^{\pm0}$ & $0^{\pm0}$ \\
    queen11\_11 & 121 & 11 & \hspace{5pt}$5.9^{\pm0.07}$ & 6& 6 & \textbf{4} & \hspace{5pt}$6^{\pm0}$ & $5.4^{\pm0.25}$ \\
    anna & 138 & 11 & \hspace{5pt}$0.2^{\pm0.04}$ & 0& 0 & 0 & \hspace{5pt}$0^{\pm0}$ & $0.4^{\pm0.18}$ \\
    2-Insertions\_4 & 149 & 4 & \hspace{5pt}$1.5^{\pm0.05}$ & 1& 1 & 1 & \hspace{5pt}$1^{\pm0}$ & $1^{\pm0}$ \\
    queen13\_13 & 169 & 13 & \hspace{5pt}$6.5^{\pm0.07}$ & 10& 9 & \textbf{4} & \hspace{5pt}$8^{\pm0}$ & $6.3^{\pm0.24}$ \\
    myciel7 & 191 & 8 & \hspace{5pt}$0.6^{\pm0.06}$ & 0& 0 & 0 & \hspace{5pt}$0.3^{\pm0.14}$ & $0.2^{\pm0.11}$ \\
    homer & 561 & 13 & \hspace{5pt}$1^{\pm0.07}$ & 0& 0 & 0 & \hspace{5pt}$0^{\pm0}$ & $0.6^{\pm0.22}$ \\
    \hline
    Average &  &  & \hspace{5pt}$1.9^{\pm0.01}$ & 1.95& 2 & 1.2 & \hspace{5pt}$1.92^{\pm0.05}$ & $1.35^{\pm0.05}$ \Tstrut\\
    \hline
    \end{tabular}}
\end{center}
\label{table:lemos_comparisons}
\caption{Comparison of ReLCol with other construction algorithms on graphs from the COLOR02 benchmark dataset. Values indicate how many more colours are required than the chromatic number, $\chi$. For each graph, Random is run 100 times and the average and standard error are reported. For Gianinazzi et al. and ReLCol, the 12 heuristics are run and the average and standard error are reported. A bold number indicates that an algorithm has found the unique best colouring amongst the algorithms.}
\end{table}

\subsection{A class of graphs on which ReLCol outperforms DSATUR}

\begin{figure}[h]
\begin{center}
\includegraphics[width=7.25cm]{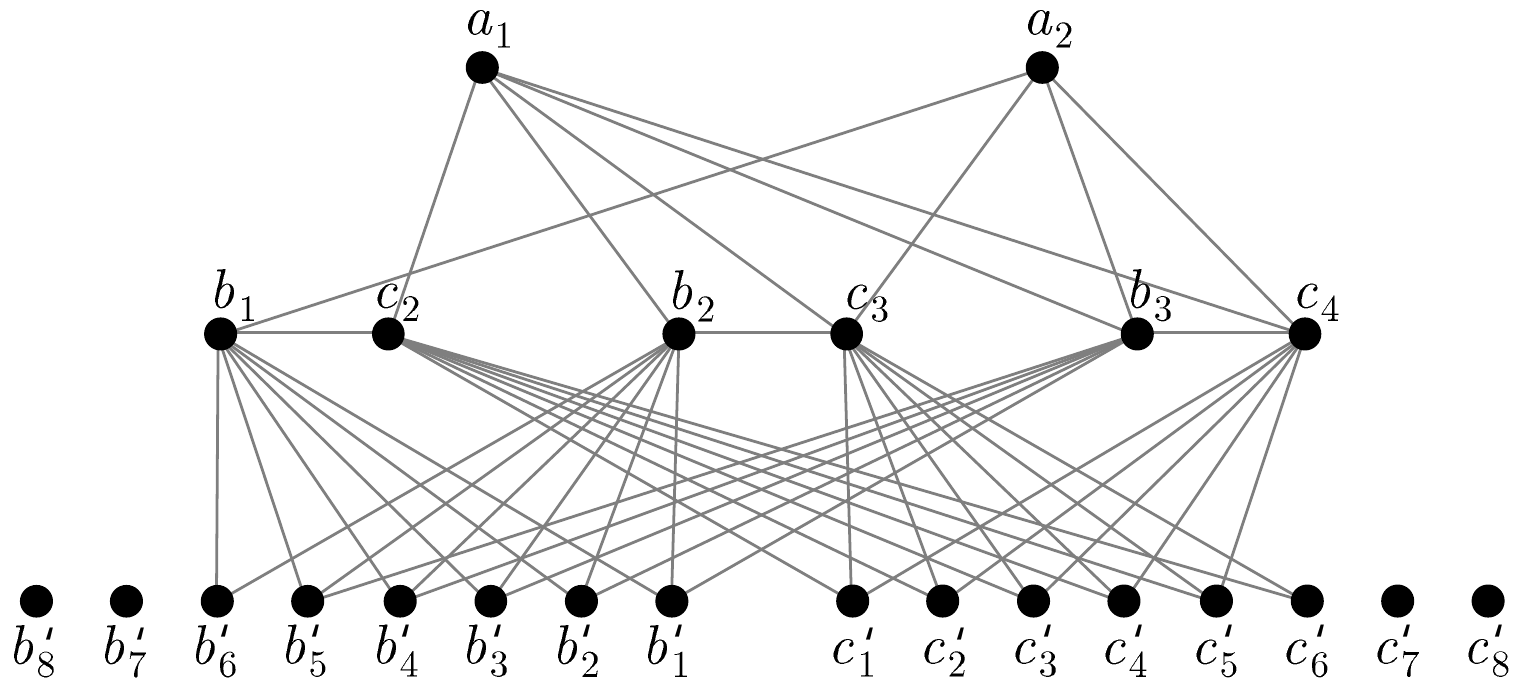}
\caption{Example of a Spinrad graph on 24 vertices, generated with $m=4$.}
\label{fig:HC_dsatur_graph}
\end{center}
\end{figure}

Although the previous results suggest that ReLCol does not outperform DSATUR on general graphs, there do exist classes of graphs on which DSATUR is known to perform poorly. One such class is presented in \cite{spinrad1985worst}; these graphs, which we refer to as \textit{Spinrad graphs}, are constructed as follows:
\begin{enumerate}
    \item Fix the number of vertices $n$ such that $n \pmod 7 = 3$, and let $m = \frac{n+4}{7}$. 
    \item Partition the vertices into 5 disjoint sets as follows:
    \begin{align*}
        &A = \{a_1, a_2, \cdots, a_{m-2}\} &\hspace{.25cm}&B = \{b_1, b_2, \cdots, b_{m-1}\} &\hspace{.25cm}&C = \{c_2, c_3, \cdots, c_m\} \\
        & &\hspace{.25cm}&B^{\prime} = \{b^{\prime}_1, b^{\prime}_2, \cdots, b^{\prime}_{2m}\} &\hspace{.25cm}&C^{\prime} = \{c^{\prime}_1, c^{\prime}_2, \cdots, c^{\prime}_{2m}\}
    \end{align*}
    \item Add the following sets of edges:
    \begin{align*}
        & E_{A}^{B} = \{ (a_i,b_j) {:\ } i \neq j\} &    & E_{A}^{C} = \{ (a_i,c_j) {:\ } i < j\} &    & E_{B}^{C} = \{ (b_{i-1},c_{i}) {:\ } 2 < i < m\} 
    \end{align*}
    Plus:
    \begin{itemize}
        \item $\forall b \in B$, add edges to vertices in $B'$ such that the degree of $b$ is $2m$.
        \item $\forall c \in C$, add edges to vertices in $C'$ such that the degree of $c$ is $2m$.
    \end{itemize}
\end{enumerate}
An example of such a graph with $m=4$ is shown in Fig. \ref{fig:HC_dsatur_graph}. Note that some of the vertices in $B^{\prime}$ and $C^{\prime}$ may be disconnected; they exist simply to ensure that the vertices in $B$ and $C$ all have degree $2m$. 

The vertices can be partitioned into 3 disjoint sets $A \cup B^{\prime} \cup C^{\prime}$, $B$ and $C$. Given that there are no edges between pairs of vertices in the same set, the chromatic number of the graph is 3. However the DSATUR algorithm assigns the same colour to vertices $a_i$, $b_i$ and $c_i$, meaning it uses $m$ colours for the whole graph. A proof of this is provided in \cite{spinrad1985worst}.

\begin{figure}[h]
\begin{center}
\includegraphics[trim={2.5cm 0.5cm 2.5cm 1cm}, width=12cm]{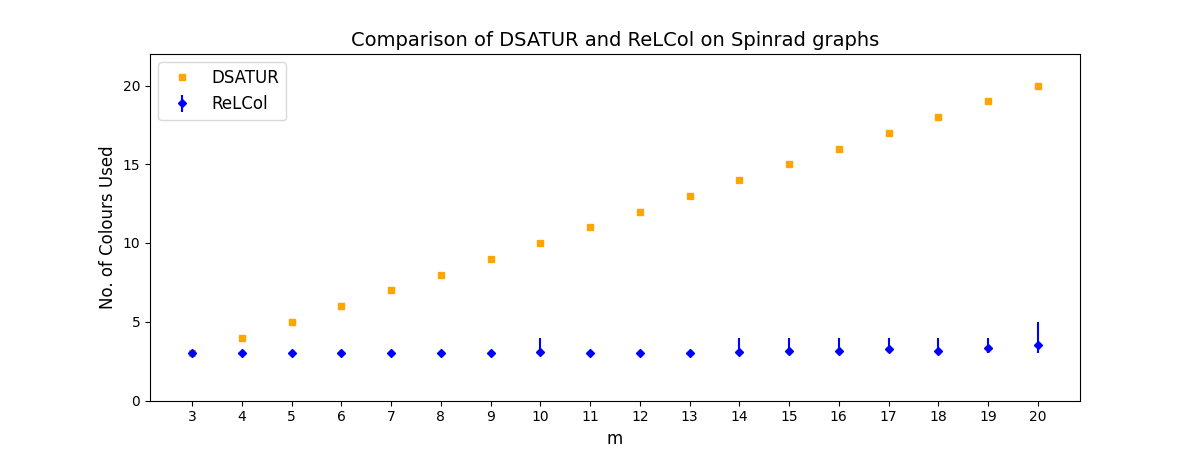}
\caption{ReLCol outperforms DSATUR on Spinrad graphs. Error bars show the maximum and minimum number of colours used by the 12 ReLCol-generated heuristics.}
\label{fig:spinrad_comparison}
\end{center}
\end{figure}

Fig. \ref{fig:spinrad_comparison} shows that ReLCol vastly outperforms DSATUR on Spinrad graphs, in most cases identifying an optimal colouring using the minimum number of colours. This indicates that despite their similar performance on general graphs, ReLCol has learned a heuristic that is selecting vertices differently to DSATUR.

\subsection{Scalability of ReLCol}
While we have shown that ReLCol is competitive with existing construction heuristics, and can outperform them on certain graph classes, our results suggest that the ability of ReLCol to colour general graphs effectively may reduce when the test graphs are significantly larger than those used for training.

\begin{figure}[h]
\begin{center}
\includegraphics[trim={2.5cm 0.5cm 2.5cm 1cm}, width=12cm]{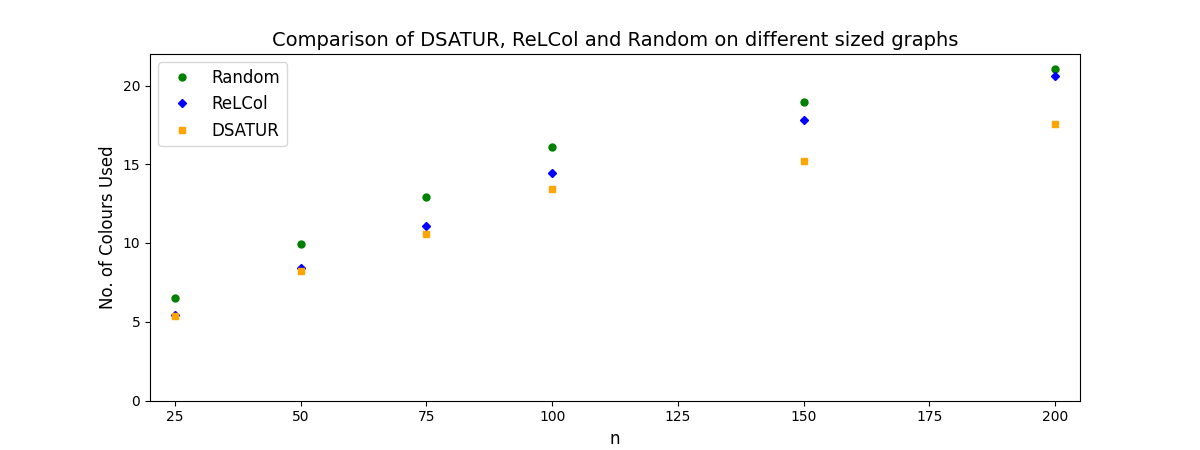}
\caption{As the graph size increases, the performance of ReLCol moves from being similar to DSATUR towards Random. This suggests that there are limitations to how well ReLCol generalises to graphs larger than those seen during training.}
\label{fig:larger_graphs}
\end{center}
\end{figure}

This can be observed in Fig.~\ref{fig:larger_graphs}, which compares the performance of DSATUR, ReLCol, and Random on graphs of particular sizes generated using the same process as our training dataset. The degradation in performance could be a result of the nature of the training dataset, whose constituent graphs have no more than 50 vertices: for graphs of this size DSATUR and ReLCol seem to achieve comparable results,
and much better than Random, but the performance of ReLCol moves away from DSATUR towards Random as the graphs grow in size.


\subsection{Representing the state as a complete graph} \label{section:complete_graph_results}
\begin{figure}[h]
\begin{center}
\includegraphics[trim={2.5cm 0.5cm 2.5cm 1cm}, width=12cm]{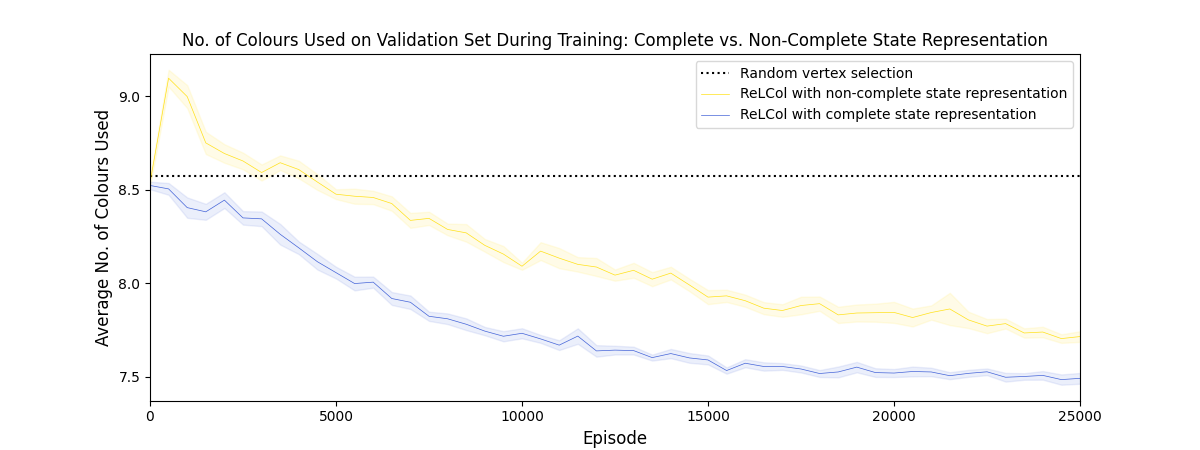}
\caption{Our complete graph representation results in faster learning and better final performance compared to the standard GNN representation.}
\label{fig:complete_vs_non_complete}
\end{center}
\end{figure}

To demonstrate the benefit of the proposed complete graph representation, 
we compare ReLCol with a version that preserves the topology of the original graph, meaning that within the GNN, messages are only passed between pairs of adjacent vertices. Fig.~\ref{fig:complete_vs_non_complete} compares the number of colours used by each version when applied to a validation dataset periodically during training. The validation dataset is composed of 100 graphs generated by the same mechanism as the training dataset. The complete graph representation clearly leads to faster learning and significantly better final performance.

\section{Conclusions}
We have proposed ReLCol, a reinforcement learning algorithm based on graph neural networks that is able to learn a greedy construction heuristic for GCP. The ReLCol heuristic is competitive with DSATUR, a leading greedy algorithm from the literature, and better than several other comparable methods. We have demonstrated that part of this success is due to a novel (to the best of our knowledge) complete graph representation of the graph within the GNN. Since our complete graph representation seems to perform much better than the standard GNN representation, we intend to investigate its effect in further RL tasks with graph-structured data. We also plan to incorporate techniques for generalisability from the machine learning literature to improve the performance of the ReLCol heuristic on graphs much larger than the training set. 

An advantage of automatically generated heuristics is that they can be tuned to specific classes of problem instances by amending the training data, so exploring the potential of ReLCol to learn an algorithm tailored to specific graph types would be an interesting direction. Finally, given that the ReLCol heuristic appears to work quite differently from DSATUR, further analysis of how it selects vertices may yield insights into previously unknown methods for GCP.

\section*{Acknowledgements}
G. Watkins acknowledges support from EPSRC under grant EP/L015374/1.\\
G. Montana acknowledges support from EPSRC under grant EP/V024868/1.\\
We thank L. Gianinazzi for sharing the code for the method presented in \cite{gianinazzi2021learning}.

%
%
%
%

\bibliographystyle{splncs04}
\bibliography{references}

\end{document}